\documentclass[conference]{IEEEtran}
\IEEEoverridecommandlockouts
\usepackage{cite}
\usepackage{amsmath,amssymb,amsfonts}
\usepackage{algorithmic}
\usepackage{graphicx}
\usepackage{textcomp}
\usepackage{xcolor}
\usepackage{epsfig}
\usepackage{amsthm}
\usepackage{multicol}
\usepackage{calc}
\usepackage{amstext}
\usepackage{float}
\def\BibTeX{{\rm B\kern-.05em{\sc i\kern-.025em b}\kern-.08em
    T\kern-.1667em\lower.7ex\hbox{E}\kern-.125emX}}

\makeatletter
\newcommand{\linebreakand}{%
    \end{@IEEEauthorhalign}
    \hfill\mbox{}\par
    \mbox{}\hfill\begin{@IEEEauthorhalign}
}

\begin{document}
\title{Unsupervised Thematic Clustering of \emph{\emph{hadith}} Texts
using the Apriori Algorithm\\}
%\thanks{PPL_B Kelompok 1}

\author{
\IEEEauthorblockN{1\textsuperscript{st} Wisnu Uriawan}
\IEEEauthorblockA{\textit{Informatics Department}\\
\textit{UIN Sunan Gunung Djati Bandung}\\
Jawa Barat, Indonesia\\
wisnu\_u@uinsgd.ac.id}
\and
\IEEEauthorblockN{2\textsuperscript{nd} Achmad Ajie Priyajie}
\IEEEauthorblockA{\textit{Informatics Department}\\
\textit{UIN Sunan Gunung Djati Bandung}\\
Jawa Barat, Indonesia\\
achmadajie74@gmail.com}
\and
\IEEEauthorblockN{3\textsuperscript{rd} Angga Gustian}
\IEEEauthorblockA{\textit{Informatics Department}\\
\textit{UIN Sunan Gunung Djati Bandung}\\
Jawa Barat, Indonesia\\
diazzangga3@gmail.com}
\linebreakand
\IEEEauthorblockN{4\textsuperscript{th} Fikri Nur Hid\emph{ayat}}
\IEEEauthorblockA{\textit{Informatics Department}\\
\textit{UIN Sunan Gunung Djati Bandung}\\
Jawa Barat, Indonesia\\
fkrnhdyt@gmail.com}
\and
\IEEEauthorblockN{5\textsuperscript{th} Muhamad Fikri Zaelani}
\IEEEauthorblockA{\textit{Informatics Department}\\
\textit{UIN Sunan Gunung Djati Bandung}\\
Jawa Barat, Indonesia\\
fikrimuhamadzael@gmail.com}
\and
\IEEEauthorblockN{6\textsuperscript{th} Sendi Ahmad Rafiudin}
\IEEEauthorblockA{\textit{Informatics Department}\\
\textit{UIN Sunan Gunung Djati Bandung}\\
Jawa Barat, Indonesia\\
sendymurphy@gmail.com}
}

\maketitle

\begin{abstract}
This research stems from the urgency to automate the thematic grouping of \emph{\emph{hadith}} in line with the growing digitalization of Islamic texts. Based on a literature review, the unsupervised learning approach with the Apriori algorithm has proven effective in identifying association patterns and semantic relations in unlabeled text data. The dataset used is the Indonesian Translation of the \emph{\emph{hadith}} of Bukhari, which first goes through preprocessing stages including case folding, punctuation cleaning, tokenization, stopword removal, and stemming. Next, an association rule mining analysis was conducted using the Apriori algorithm with support, confidence, and lift parameters. The results show the existence of meaningful association patterns such as the relationship between ``\emph{rakaat}-prayer'', ``verse-revelation'', and ``\emph{\emph{hadith}}-story'', which describe the themes of worship, revelation, and \emph{\emph{hadith}} narration. These findings demonstrate that the Apriori algorithm has the ability to automatically uncover latent semantic relationships, while contributing to the development of digital Islamic studies and technology-based learning systems.

\end{abstract}

\begin{IEEEkeywords}

Apriori, unsupervised learning, text mining, hadis, asosiasi tematik.

\end{IEEEkeywords}

\section{Introduction} \label{sec:introduction}

\emph{\emph{hadith}} is one of the main sources of Islamic teachings after the Qur'an which contains theological, historical and moral values \cite{inproceedings}.With the advancement of the digital era, various efforts have been made to transform \emph{\emph{hadith}} books into digital form to facilitate access, searching, and scientific study. However, this digitization process presents new challenges in terms of managing, grouping, interpreting, and classifying \emph{\emph{hadith}} by theme. The ever-increasing number of \emph{\emph{hadith}} digitally, both in text form and in online databases, raises the need for new approaches capable of automatically identifying the interconnectedness of meaning and context.

Traditionally, the determination of themes in \emph{\emph{hadith}} has been done manually by \emph{\emph{hadith}} scholars using philological and contextual approaches. While accurate, this approach is time-consuming and prone to interpreter subjectivity. At the same time, the rapid development of information technology has opened up opportunities for the application of machine learning and text mining in the management of religious texts. In this context, the use of unsupervised learning approaches is highly relevant, as they are able to recognize thematic patterns and structures from text data without the need for initial labels or annotations\cite{ramponi2020unsupervised}. This approach is in line with the spirit of automation in digital Islamic studies which aims to expand access to religious knowledge while increasing the objectivity of analysis of \emph{\emph{hadith}} texts.

One of the algorithms widely used in the unsupervised learning domain is Apriori, which is widely implemented in association rule mining to find relationships between items or words in a data set \cite{alfianzah2020apriori} \cite{hibnastiar2025apriori}. In the context of \emph{\emph{hadith}} texts, this algorithm can be used to extract semantic relationships between frequently occurring words or concepts, thus creating a thematic structure that reflects the latent meaning behind the \emph{\emph{hadith}} narrative. Thus, analysis using Apriori is not simply a frequency count but also an attempt to build a data-driven conceptual understanding of religious texts.

The application of text mining and natural language processing (NLP) in the analysis of Islamic texts can help discover previously hidden patterns of knowledge, as well as support the birth of more systematic and measurable data-based thematic interpretations\cite{Doskarayev2023}.The use of this method also strengthens the Digital Islamic Humanities initiative which seeks to integrate computational methodologies in the study of classical texts to maintain their relevance in the digital age \cite{FatinaWahid2021}.

Furthermore, this research supports the achievement of Sustainable Development Goal (SDG) 4: Quality Education, particularly in encouraging access to inclusive and technology-based education \cite{do2020humanrights} \cite{ferguson2020sdg} \cite{sayed2020quality} \cite{franco2020sdg}. By providing an automated method for understanding the content of \emph{\emph{hadith}}, this approach expands access to quality learning for students, researchers, and the general public, while bridging traditional values with modern technology within the framework of Islamic education.

This research also contributes to the rapidly developing field of Artificial Intelligence for Islamic Studies (AIIS) in the last decade. AIIS seeks to combine artificial intelligence approaches with Islamic texts to extract meaning, analyze \emph{sanad} (chain of transmission), and identify conceptual relationships between \emph{\emph{hadith}}. As explained by Al-Hasan et al. (2021), unsupervised clustering and association rule mining methods can be used to uncover the semantic dimensions of \emph{\emph{hadith}} texts, which are often missed through manual analysis \cite{Alhasan2021}.Therefore, the integration of the Apriori algorithm in \emph{\emph{hadith}} studies is not only a technical innovation, but also a methodological transformation in understanding the sources of Islamic teachings more broadly, deeply, and objectively.

Based on this urgency, this study proposes the application of the Apriori algorithm within an unsupervised learning framework to extract word association patterns in digital Indonesian-language \emph{\emph{hadith}} texts. This approach is expected to automatically form thematic groupings of \emph{\emph{hadith}}, facilitate the process of discovering new meanings, and strengthen the direction of developing an artificial intelligence-based religious information system that supports the vision of digital Islamic education and quality education.

\section{Related Work} \label{sec:related-work}

Previous research has demonstrated the significant potential of unsupervised learning approaches in detecting patterns from unlabeled data. One study developed an AI-based framework capable of recognizing hidden relationships among crime data, with results demonstrating the model's effectiveness in clustering incidents based on pattern similarities. Although the context of this research is in the realm of crime analysis, the application of unsupervised methods and association patterns from text data demonstrates high relevance for thematic clustering of \emph{\emph{hadith}}, which are also unlabeled and require mapping the meaning of latent patterns \cite{raja2021ai}.

A comparative analysis of two popular algorithms, K-Means for clustering and Apriori for association patterns, shows that K-Means excels in forming homogeneous clusters, while Apriori is more effective in finding relationships between items within a cluster. This approach demonstrates that the combination of clustering techniques and association algorithms can enrich the grouping of \emph{\emph{hadith}} texts by providing additional insight into the semantic relationships within them \cite{dharshinni2019kmeans}.

Another study demonstrated the application of the Apriori algorithm in analyzing sales transactions to aid inventory management decisions. The results were concrete association rules, such as the relationship between purchases of certain items based on consumer habits. Although this research was conducted in a retail context, the approach is relevant because it illustrates Apriori's ability to extract meaningful rules from raw data. This ability is also needed to discover thematic relationships between the contents of \emph{\emph{hadith}} texts \cite{tarigan2022implementasi}.

Another study examined the process of clustering text-based documents through feature extraction and classification using the Naïve Bayes method. Although this approach includes supervised learning, the study emphasized the importance of preprocessing stages such as tokenization and feature selection to improve document clustering accuracy. Its relevance to clustering \emph{\emph{hadith}} texts lies in the importance of preprocessing the text as a key to success in unsupervised approaches such as clustering and association algorithms \cite{nurmalasari2022implementasi}.

Another unsupervised learning approach demonstrated the effectiveness of the K-Means algorithm in forming homogeneous clusters from various types of unlabeled data. The results of this study confirmed that K-Means is capable of efficiently identifying latent structures in data, provided the number of clusters is selected appropriately. This aligns with efforts to cluster \emph{\emph{hadith}} texts based on similar themes, where K-Means can be used to automatically map documents into thematic clusters \cite{sinaga2020unsupervised}.

Another study introduced an algorithm for automatically generating concept maps from text using text analysis and association rule mining techniques. One of the main techniques is the Apriori algorithm, which is used to extract relationships between concepts based on word frequency and association. The results of this study are concept maps that automatically represent the knowledge structure within documents, demonstrating Apriori's great potential in constructing meaningful semantic representations from text collections \cite{shao2020conceptmap}.

A similar approach was also applied to identify sales transaction patterns for local brands, where the Apriori algorithm was able to transform sales data into strategic information for marketing decision-making. The results showed that the association patterns discovered can support the development of data-driven strategies, relevant to the goal of clustering \emph{\emph{hadith}} texts, which also focuses on discovering meaningful relationships between concepts \cite{az2024implementasi}.

Another study utilized the Apriori algorithm to provide book placement recommendations based on borrowing patterns. The resulting system was able to identify relationships between frequently searched-for book categories, thereby improving management and search efficiency. This approach can be used as an analogy for thematic mapping of \emph{\emph{hadith}} texts based on the co-occurrence of words or concepts within them \cite{yudhatama2024penerapan}.

Research combining the K-Means and Apriori methods shows that the combination is capable of forming representative clusters and identifying strong relationships between items with significant support and confidence values. This integration of clustering and association rule techniques is crucial for exploring the latent structure and semantic relationships of data, including in the context of \emph{\emph{hadith}} texts \cite{alinafiah2024implementasi}.

The study demonstrates the application of the Apriori algorithm in the retail sector to analyze inter-product relationship patterns based on large sales data sets. Using an association rule mining approach, the study was able to identify product combinations with high support and confidence values, resulting in association rules that can be used for promotional strategies and product arrangement in stores. The relevance of this study to the study of Unsupervised Thematic Clustering of \emph{\emph{hadith}} Texts using the Apriori Algorithm lies in the ability of the same algorithm to discover latent relationships between items or concepts from large data sets \cite{7577586}. In the context of \emph{\emph{hadith}} texts, this approach can be utilized to uncover associations of meaning between terms or themes, thus assisting in more meaningful and data-based thematic grouping.

Finally, the Apriori algorithm was also applied to identify spare parts purchasing patterns by generating high-confidence association rules. The results show that Apriori can aid efficient product layout and support strategic decisions. This application conceptually parallels the thematic clustering of \emph{\emph{hadith}} texts, which aims to discover hidden relationships between concepts or terms within the text \cite{nurhidayanti2022apriori}.

Many issues arise in the study of \emph{\emph{hadith}} that are trending in the discipline, ranging from the digitization of \emph{\emph{hadith}} data to case-based analyses concerning approximate narrator chains. However, this paper does not focus on estimating, justifying, or authenticating \emph{\emph{hadith}}; instead, it highlights the application of data mining techniques on \emph{\emph{hadith}} datasets. In this study, the \emph{\emph{hadith}} dataset was incorporated into a machine learning tool, specifically text clustering, to examine the potential clustering of \emph{\emph{hadith}} texts based on frequently appearing words. The objective was to apply BERT modeling to cluster \emph{\emph{hadith}} from Sahih Bukhari and Muslim, exploring the feasibility of using NLP techniques for this task. The findings of this research contribute to the advancement of more sophisticated and accurate methods for text analysis \cite{10903699}.

Based on various previous studies, it appears that the Apriori algorithm and clustering methods such as K-Means have proven effective in uncovering hidden patterns in various types of data, including transaction data, inventory, and text documents. The use of an unsupervised learning approach provides flexibility in analyzing data without the need for labels, making it highly suitable for thematic clustering of \emph{\emph{hadith}} texts, which naturally lack theme annotations. The integration of Apriori and clustering opens up significant opportunities for extracting knowledge structures and semantic relationships between concepts in religious texts. Therefore, this study seeks to build on the contributions of previous studies by applying a similar approach specifically to \emph{\emph{hadith}} texts for the purpose of automatic and systematic interpretation and knowledge discovery.

\section{Methodology} \label{sec:methodology}

The increasing use of machine learning in \emph{hadith} studies not only enables clustering and thematic grouping using models such as BERT but also opens opportunities to uncover deeper semantic relationships within the texts. Beyond clustering approaches, unsupervised techniques can be employed to discover patterns that reveal how certain concepts, terms, or themes frequently appear together in classical narrations. These methods complement prior analyses by providing another perspective on textual structure and hidden associations within the \emph{hadith} corpus.

\begin{figure}[H]
\centering
\includegraphics[width=0.55\linewidth]{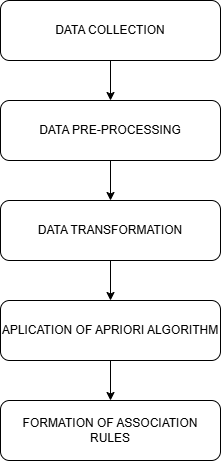}
\caption{Research Methodology}
\label{fig:metode}
\end{figure}

This research employs an unsupervised learning approach with the Apriori algorithm to extract word associations from \emph{hadith} texts. This method aims to identify thematic connections between words to establish meaningful theme groups, serving as a basis for new knowledge. The technical stages of the process are shown in Figure \ref{fig:metode}.

\subsection{Data Collection}

The \emph{hadith} dataset used in this study was sourced from the 'Indonesian Translation of the \emph{hadith} of Bukhari's which is available on Dataverse. This dataset comprises thousands of \emph{hadith} text entries in Indonesian (translated), complete with their corresponding \emph{hadith} ID.

\begin{enumerate}
    \item Amount of Data
    ± 6000 entry (select a subset if necessary)
    \item Data Format
    Excel (.xlsx)
    \item Data Source
    Indonesian Translation of the \emph{hadith} of Bukhari (Single-label)\cite{FK2/GWSEWB_2022}
\end{enumerate}{Data Format}

\begin{table}[H]
\centering
\caption{dataset attributes}
\label{tab:atribut_dataset}
\begin{tabular}{|p{0.10\textwidth}|p{0.35\textwidth}|}
\hline
\centering\textbf{dataset attributes} & \centering\textbf{Information} \tabularnewline
\hline
No & Sequential number for each \emph{hadith} entry in the dataset, used as a unique identifier or data index. \\
\hline
BAB & Indicates the chapter number (sub-category/topic) to which the \emph{hadith} belongs. \\
\hline
Hadist & The textual content of the \emph{hadith}. This is the main component analyzed in the research, particularly in the \textit{pre-processing} stage and association analysis. \\
\hline
Len & The length of the \emph{hadith} text in characters. This can be used for additional analysis such as text-length filtering or statistical distribution of \emph{hadith} lengths. \\
\hline
\end{tabular}
\end{table}

\subsection{Pre-processing Data}
The objective of the preprocessing stage is to clean and prepare the text so it can serve as transactions for association rule mining. The procedures performed are as follows:

\begin{figure}[H]
\centering
\includegraphics[width=0.4\linewidth]{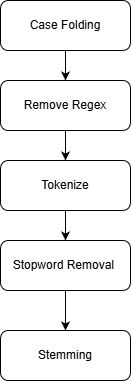}
\caption{Pre-Prossesing}
\label{fig:pre-prossesing}
\end{figure}

\begin{enumerate}
\item Case Folding 

Case folding is the process of converting all characters in the text into (lowercase) \cite{lischner2014case}. The objective is to standardize the word format. This is necessary because in text processing, terms like ''Islam`` ''islam`` and ''ISLAM`` would be treated as three distinct entities if normalization is not applied. Therefore, every line of the \emph{hadith} text is converted to lowercase, ensuring that words appearing in capitalized forms are still processed as the same word.

\item Punctuation Removal

Punctuation and non-alphabetic characters (such as periods, commas, exclamation marks, numbers, and other symbols) lack semantic contribution in the word association formation process. Consequently, these characters are removed from the text.

\item Tokenization (Lexical Analysis)

Tokenization is the process of splitting the text into word units called tokens \cite{rosid2020improving}.This step is crucial for individually analyzing words and forming word transactions. To achieve this, each line of text is broken down based on spaces, yielding a list of words (tokens) that can be analyzed separately.

\item Stopword Removal

Stopwords are common words that appear frequently but lack significant semantic importance in thematic analysis, such as ''\emph{dan}`` (and), ''\emph{yang}`` (which), ''\emph{dari}`` (from), and so on \cite{rosid2020improving}. n the \emph{hadith} context, there are also religious terms that occur too frequently yet lack relevance in forming thematic clusters, such as ''\emph{rasulullah}`` (Messenger of God), ''\emph{nabi}`` (Prophet), ''\emph{shallallahu}`` (may God bless him and grant him peace), and ''\emph{ibnu}`` (son of). The objective of this stage is to filter out words that do not directly contribute to the formation of thematic associations within the \emph{hadith} texts. By eliminating these terms, the subsequent analysis can focus more effectively on the core meaning of each \emph{hadith} narration, using the following procedure:
\begin{enumerate} 
\item Retrieve the Indonesian stopword list from the NLTK library.
\item Add custom stopwords that frequently appear in \emph{hadith} texts but are not relevant for association analysis.
\item  Remove all tokens included in the stopword list.
\end{enumerate}

\item Stemming

Stemming is the process of returning a word to its base form (root form) \cite{bougar2019addressing}. For example, the words ''\emph{beriman}`` ''\emph{berimanlah}`` and ''\emph{keimanan}`` will be reduced to the root form ''\emph{iman}`` The goal is to unify various derived forms of a word so they are counted as the same entity in the association process.

\item Data Preprocessing Integration

All of the steps above are then integrated into a single sequential process and applied to each text entry in the dataset. The final result of this stage is a representation of each \emph{hadith} in the form of a clean and consistent list of root words, ready to be used for word-association analysis. This representation is referred to as a transaction, which serves as the input for data transformation and the application of the Apriori algorithm.

\item Data Transformation

After the preprocessing stage is completed, the text data in the form of lists of root-word tokens from each \emph{hadith} is then converted into a transaction format for association processing. The data transformation is carried out so that the data can be interpreted as a binary representation, indicating whether a word appears (1) or does not appear (0) in each \emph{hadith}.

This binary representation is known as one-hot encoding, which is commonly used in data mining techniques. Each \emph{hadith} is represented as a row in a table, while the columns correspond to the unique words that appear in the entire dataset. If a word appears in the \emph{hadith}, its column is assigned a value of 1, and if it does not appear, it is assigned a value of 0.

This transformation is important because the Apriori algorithm requires input in the form of binary transactions to calculate the frequency of item (word) combinations across the entire dataset. With this format, each \emph{hadith} is treated as a transaction basket containing a set of items in the form of words.

This process enables association analysis to be carried out efficiently, as the uniform data structure allows the system to easily calculate the support, confidence, and lift values for each word combination. In the context of \emph{hadith} research, this approach helps reveal relationships between religious concepts that are hidden within the narrative text. Each pair or group of words that frequently co-occurs in certain \emph{hadith}s may indicate semantic relationships that reflect specific themes, topics, or contexts of religious events.

The binary transaction representation also provides flexibility for integrating other algorithms in later analysis stages, such as clustering or thematic classification. Once the association patterns are obtained, the results can be used to group \emph{hadith}s based on similarities in their word structures. In this way, the Apriori method not only functions to discover relationships between words but also serves as a foundation for grouping \emph{hadith}s according to similar content. This approach expands the scope of \emph{hadith} text analysis from simple word exploration to a more systematic mapping of knowledge.

In its application, constructing the binary representation requires an accurate word-normalization stage to ensure that word meanings are not distorted. Each token must undergo lemmatization or stemming to guarantee that the word forms used are consistent. For example, the words ''\emph{\emph{shalat}}`` ''\emph{menunaikan \emph{shalat}}'' and ''\emph{\emph{shalat}lah}`` must be reduced to the same root form, namely ''\emph{\emph{shalat}}`` This uniformity is essential so that word frequency truly reflects conceptual connections rather than mere morphological variations.

The results of this transformation also play a role in ensuring that recurring patterns can be identified objectively. With a standardized data structure, the system can calculate word combinations with measurable complexity, free from human interpretive bias. In the context of religious text studies, this helps produce more neutral, data-driven analyses while strengthening transparency in the process of knowledge discovery.

The application of binary transaction formatting in \emph{hadith} studies reflects the integration of modern scientific approaches with classical Islamic literacy traditions. Textual and narrative data can now be processed in a quantitative form, allowing researchers to uncover regularities and thematic patterns computationally. Thus, this approach not only transforms how \emph{hadith} texts are analyzed but also broadens the horizon of Islamic scholarship toward a more empirical, adaptive, and technologically relevant direction. 
\end{enumerate}

\subsection{application of the Apriori algorithm}
Apriori is an algorithm used to discover association patterns from transactional data. In the context of this study, the algorithm is used to identify frequent itemsets, which are groups of words that often appear together within the same \emph{hadith}.

The first step in this process is determining the minimum support value, which serves as the threshold for how frequently a word combination must occur to be considered significant. For example, if the minimum support is set to 0.02, then only word combinations that appear in at least 2 percent of all \emph{hadith}s will be taken into account.

After that, the algorithm generates word combinations of increasing sizes—single words (1-itemsets), pairs of words (2-itemsets), triplets (3-itemsets), and so on. Each combination is evaluated against the support threshold. If a combination does not meet the requirement, all combinations that include it are also discarded. This makes the Apriori algorithm efficient because it uses the principle of antimonotonicity: if an itemset is not frequent, then any of its supersets will definitely not be frequent.

In its application to \emph{hadith} texts, each word that has gone through preprocessing is treated as an item, and each \emph{hadith} is treated as a transaction. The process of finding frequent itemsets is carried out iteratively to identify word combinations that commonly appear together in a single \emph{hadith}. For example, if the words ''\emph{shalat}`` and ''\emph{rakaat}`` frequently appear close to each other in several \emph{hadith}s, then their combination will be detected as a frequent 2-itemset. The higher the support value of the combination, the stronger the indication that the two words share a significant thematic relationship within the context of worship.

The frequent itemsets identified at this stage become the foundation for forming association rules. Each combination of words that meets the minimum support value will be paired in the form of $antecedent \rightarrow consequent$ to be further analyzed using the confidence and lift metrics. Thus, this stage is not only intended to calculate frequency but also to select word combinations that are suitable to serve as the basis for thematic inference. This approach allows the identification of semantic relationships to be carried out objectively based on the distribution of words across the entire \emph{hadith} corpus, without subjective intervention from the researcher.

The application of the Apriori algorithm in religious text research, such as the study of \emph{hadith}, demonstrates that computational methods can be used to uncover conceptual patterns within classical texts. The results obtained are not merely numbers or statistics, but also new insights into the connections between concepts that hold theological relevance. By combining quantitative approaches with contextual interpretation, this algorithm serves as a scientific tool that strengthens traditional qualitative analysis methods. Such an approach affirms that integrating data science with Islamic studies can produce new forms of understanding that are systematic, transparent, and profound in exploring the intellectual heritage of Islam.

\subsection{formation of association rules}
The final step in this research method is to construct association rules based on the frequent itemsets that have been identified. Association rules express an implicative relationship between two groups of words, for example, ''if word A appears, then word B is also likely to appear.`` In the context of this study, association rules are used to identify thematic relationships between words in \emph{hadith} texts. Each rule is evaluated based on three main metrics:\\

\begin{enumerate}
    \item Support:Measuring how frequently the word combination appears in the dataset.
    \item Confidence: Measuring the probability that the consequent word appears when the antecedent word appears.
    \item Lift: Measuring the strength of the relationship between the two groups of words compared to their random occurrence. A lift value greater than 1 indicates a positive association.
\end{enumerate}

Association rules that meet the confidence and lift thresholds are then analyzed to identify thematically meaningful patterns. The results of this process can be used as a foundation for forming thematic groups within \emph{hadith}s, discovering semantic relationships between words, and building \emph{hadith} content recommendation systems based on relevant topics.

The process of forming association rules is not only mechanical but also requires conceptual interpretation so that the results can be understood within the context of Islamic scholarship. Each word pair with a high lift value must be analyzed further to ensure its semantic and contextual relevance within the \emph{hadith} text. For example, the relationship between the words ''\emph{ayat}`` and ''\emph{turun}`` may statistically indicate a strong connection, but theologically, it also reinforces the theme of revelation, which is central to prophetic experience. Thus, the interpretation stage becomes an important part of ensuring that the analytical results remain aligned with the religious meanings embedded in the text.

The use of association rules in \emph{hadith} research also opens possibilities for exploring cross-thematic patterns. Combinations of words involving elements of worship, ethics, or prophetic history can be analyzed together to identify conceptual relationships across different domains. This approach helps build a more comprehensive understanding of the thematic structure of \emph{hadith}, where a concept does not stand alone but is interconnected within a broader network of meanings. Supported by metrics such as support and confidence, these analytical results can be mapped into a comprehensive, data-driven model of thematic relationships.

The validated association rules can serve as the foundation for developing digital Islamic knowledge systems. Such systems can support automatic classification, semantic search, and theme-based \emph{hadith} learning. By leveraging accurate word association data, users can explore \emph{hadith}s not only through specific keywords but also through conceptual relationships mapped by the system. This approach marks an important step toward the digitalization of \emph{hadith} studies, focusing not only on preserving the text but also on uncovering deeper meanings and semantic relationships among religious concepts.

\section{Result and Discussion} \label{sec:result}

\subsection{Result}
\subsubsection{Thematic Clustering Results}
The application of the Apriori algorithm to the collection of \emph{hadith} texts aims to extract word association patterns that indicate dominant narrative themes. With a minimum support parameter of 0.02 and a minimum lift of 0.6, the system successfully identified several strong and semantically meaningful association rules. Apriori produced relations in the form of antecedent → consequent pairs, which were then interpreted as thematic groupings based on the co-occurrence of words within \emph{hadith} documents.

Tabel \ref{tab:association_rule} The following presents ten association rules with the highest confidence values, indicating the probability that the consequent appears when the antecedent has appeared within a single transaction (\emph{hadith} document).
\begin{table*}[htb]
\centering
\scriptsize
\caption{Association Rule Mining Results}
\label{tab:association_rule}
\begin{tabular}{|c|c|c|c|c|c|c|c|}
\hline
\textbf{No} & \textbf{Antecedents} & \textbf{Consequent} & \textbf{Antecedent Support} & \textbf{Consequent Support} & \textbf{Support} & \textbf{Confidence} & \textbf{Lift} \\
\hline
1 & \emph{\emph{rakaat}} & \emph{\emph{shalat}} & 0.028543 & 0.140003 & 0.025689 & 0.900000 & 6.428440 \\
2 & \emph{laksana} & \emph{\emph{shalat}} & 0.056229 & 0.140003 & 0.033395 & 0.593909 & 4.242118 \\
3 & \emph{\emph{ayat}} & \emph{\emph{turun}} & 0.072642 & 0.082774 & 0.041815 & 0.575639 & 6.954309 \\
4 & \emph{\emph{turun}} & \emph{\emph{ayat}} & 0.082774 & 0.072642 & 0.041815 & 0.505172 & 6.954309 \\
5 & \emph{\emph{diri}} & \emph{\emph{shalat}} & 0.069930 & 0.140003 & 0.030541 & 0.436735 & 3.119470 \\
6 & \emph{\emph{hadits}} & \emph{\emph{cerita}} & 0.059797 & 0.081775 & 0.024404 & 0.408115 & 4.990678 \\
7 & \emph{\emph{cerita}} & \emph{\emph{kepada}} & 0.081775 & 0.100756 & 0.030113 & 0.368237 & 3.654730 \\
8 & \emph{\emph{malam}} & \emph{\emph{shalat}} & 0.056943 & 0.140003 & 0.020551 & 0.360902 & 2.577821 \\
9 & \emph{\emph{baca}} & \emph{\emph{ayat}} & 0.062794 & 0.072642 & 0.022406 & 0.356818 & 4.912033 \\
10 & \emph{\emph{ayat}} & \emph{\emph{baca}} & 0.072642 & 0.022406 & 0.022406 & 0.308448 & 4.912033 \\
\hline
\end{tabular}
 \end{table*}

Rule number (1) shows the strongest association between the words ''\emph{rakaat}`` and ''\emph{shalat}``, with a confidence value of 0.900000 and a lift of 6.428440. Conceptually, this pattern represents the most fundamental worship relationship in \emph{hadith} texts, where ''\emph{rakaat}`` is a core component of the practice of ''\emph{shalat}.`` Meanwhile, rules (2) to (5) broaden the scope of associations still related to worship contexts, such as ''laksana,`` ''\emph{diri},`` and ''\emph{malam},`` each representing expressions of command, spiritual awareness, and the time of performing prayer.

The theme of revelation is strongly reflected in rules (3) and (4), which capture the reciprocal relationship between ''\emph{ayat}`` and ''\emph{turun}.`` The highest lift values in these two rules (6.954309) indicate the consistency of the revelation concept within \emph{hadith} narratives, where the process of divine descent is explicitly recorded through the co-occurrence of these words. This association affirms that the algorithm can detect semantic patterns aligned with the theological meaning structure in Islamic tradition.

Other rule groups, such as (6) and (7), reveal patterns associated with the theme of \emph{hadith} transmission. The relationships between ''\emph{hadits},`` ''\emph{cerita},`` and ''\emph{kepada}`` reflect the oral narrative form in the transmission process, consistent with the \emph{sanad} and \emph{matan} structures commonly found in phrases such as ''telah men\emph{cerita}kan \emph{kepada} kami...`` Although their confidence values are not as high as those in the worship group, the high lift values indicate that these associations remain semantically significant and reflect distinctive characteristics of \emph{hadith} literature.

On the other hand, rules (9) and (10) highlight the connection between ''\emph{baca}`` and ''\emph{ayat},`` representing the themes of recitation worship and Qur’anic revelation. Both have consistently high lift values (4.912033), indicating a strong bidirectional relationship in the context of the command to read revelation. These findings reinforce the view that word associations can serve as important indicators for automatically identifying religious themes.

The association patterns presented in the table do not merely show repeated words but also reflect how \emph{hadith} texts construct meaning through conceptual co-occurrence. The relationships between words illustrate an interconnected meaning structure, where themes of worship, revelation, and transmission form mutually reinforcing thematic networks. These results show that the language of \emph{hadith} is not random but is semantically organized according to the structure of religious knowledge.

Interpretation of the association rules also demonstrates the capability of this approach to uncover thematic patterns relevant to classical Islamic scholarly traditions. The associations produced in an unsupervised manner remain consistent with meanings recognized in \emph{hadith} studies, such as the concepts of \emph{sanad}, \emph{matan}, and worship. This approach indicates that data-driven modeling can support the reconstruction of meaning in religious texts without losing their normative and theological context.

The association results presented in the table
\ref{tab:association_rule}indicates that the Apriori method is capable of capturing the main themes within \emph{hadith} texts, including aspects of worship, revelation, and transmission. These patterns can serve as the foundation for developing thematic annotations, data-driven \emph{hadith} ontologies, and semantic search systems that enable contextual meaning retrieval. Thus, this approach not only supports academic analysis but also becomes an important basis for developing intelligent and structured Islamic knowledge technologies.

\subsubsection{Thematic Interpretation and Knowledge Findings}

The association rules generated from the application of the Apriori algorithm reveal semantic patterns that reflect the main themes in \emph{hadith} texts. Although obtained in an unsupervised manner and without manual annotation, these rules exhibit conceptual alignment with the meaning structures commonly recognized in \emph{hadith} studies, particularly in the domains of worship, revelation, and oral transmission. Grouping the rules into these themes forms the basis for potential knowledge representations that can be developed into thematic annotations, ontologies, and knowledge-navigation systems for religious texts.

The association rules can be viewed as a form of automatic thematic pattern discovery with the potential to serve as the foundation for developing representations of religious knowledge. Grouping the rules into themes such as worship, revelation, ethics, and oral transmission can assist in constructing thematic annotation systems, scientific ontologies of \emph{hadith} studies, and semantic navigation tools for religious texts. Within the fields of computer science and digital Islamic studies, these findings open significant opportunities for building data-mining-based models of \emph{hadith} comprehension, capable of presenting inter-concept relationships in a conceptual, systematic, and structured manner without neglecting the values of traditional scholarship.

These findings demonstrate that association methods function not only as exploratory tools for discovering relationships between words but also as bridges toward the development of computational Islamic knowledge systems. By utilizing the association rules produced, a system can identify relational patterns that reflect broader religious understanding. For instance, the associations among the words ''\emph{shalat},`` ''\emph{rakaat},`` and ''\emph{malam}`` do not merely illustrate linguistic relationships but also imply specific worship practices with deep spiritual significance. This approach enables machines to understand contextual meaning in religious texts in a more semantic, rather than merely syntactic, manner.

The application of association analysis in the context of \emph{hadith} also contributes to the advancement of digital thematic exegesis. By tracing relationships between words and themes within \emph{hadith}, researchers can identify how particular concepts are consistently connected to other teachings across various textual sources. This helps construct an interpretive framework grounded in data-driven evidence, strengthening the validity of religious interpretation based on the frequency and consistency of concept occurrence. Thus, this method can serve as a scientific tool that enriches traditional approaches to understanding religious messages.

The potential applications of these findings extend beyond academic research and can be developed into educational and da'wah technologies powered by artificial intelligence. Digital systems supported by \emph{hadith} association data can be used to explore themes, answer religious questions, or provide text recommendations based on specific contexts. For example, users searching for the topic of ''revelation`` can automatically be directed to \emph{hadith} closely associated with the words ''\emph{ayat}`` ''\emph{turun}`` and ''wahyu`` In this way, the results of this research not only enrich academic discourse but also expand public access to deep, structured, and data-driven understanding of \emph{hadith}.

\begin{enumerate}
    \item Worship Theme

    he following four rules demonstrate strong relationships related to religious practices, particularly prayer \emph{shalat}:
    \begin{enumerate}
    \item (\emph{rakaat}) → (\emph{shalat}) dengan confidence 0.900000 dan lift 6.428440,
    \item (laksana)  → (\emph{shalat}) dengan confidence 0.593909 dan lift 4.242118
    \item (\emph{diri}) → (\emph{shalat}) confidence 0.436735, dan lift 3.119470
    \item (\emph{malam}) → (\emph{shalat}) (confidence 0.360902, dan lift 2.577821)
    \end{enumerate}
    The four rules reflect ritual practices in Islam. All of them emphasize aspects of performing prayer (\emph{shalat}) in various forms and conditions. The presence of the word ''\emph{malam}`` (night) in association with ''\emph{shalat}`` indicates the recurring theme of night prayers in the \emph{hadith}, while ''\emph{diri}`` (self) relates to the personal and spiritual dimension of worship.

The knowledge discovered here shows that the concept of worship in the \emph{hadith} is not conveyed solely through the explicit word ''\emph{shalat}``, but also through patterns of co-occurring meanings such as ''\emph{rakaat}``, ''\emph{malam}``, and ''\emph{diri}``. These patterns demonstrate that the interpretation of worship is constructed through temporal, spiritual, and structural contexts.

    \item Prophetic Theme

   The following two rules illustrate a narrative about the delivery of revelation that does not always explicitly mention ''revelation`` or ''angel`` but is reflected through the co-occurrence of the words ''descend`` ''verse`` and ''read``.
    \begin{enumerate}
        \item (\emph{ayat}) → (\emph{turun}) (confidence 0.575639, lift 6.954309),
        \item (\emph{baca}) → (\emph{ayat}) dan (\emph{ayat}) → (\emph{baca}) (lift 4.912033)
    \end{enumerate}

    The pairing of ''\emph{ayat}`` and ''\emph{turun}`` indicates a close reciprocal relationship in the context of revelation. This suggests that \emph{hadith} mentioning ''\emph{ayat}`` are very likely to also mention the process of their ''\emph{turun}`` referring to the divine context of revelation being sent down to the Prophet Muhammad (peace be upon him).

The associations (verse–descend) and (read–verse) reflect the narrative of revelation and the Prophet’s interaction with it. These bidirectional relations highlight the strong connection between the words ''read`` and ''verse,`` which appear in early prophetic \emph{hadith}s, particularly those related to the command to read (iqra’) and its relationship to the Qur’anic verses. This finding reflects the cognitive and spiritual processes of prophethood—how revelation is conveyed and received through the act of reading and internalizing the divine message.

Overall, these associations reveal a fundamental theme in \emph{hadith} literature: the prophetic function as the receiver, transmitter, and teacher of revelation, consistently manifested in the linguistic structure and content of the \emph{hadith} texts.

As a knowledge finding within the \emph{hadith} texts, the narrative of divine transmission can be reconstructed through semantic traces such as ''verse`` and ''descend`` or ''read,`` without the explicit mention of prophetic or angelic entities. This shows that the system is capable of detecting the prophetic process implicitly.
    
    \item Theme of \emph{hadith} Transmission

   The associations in this theme highlight how \emph{hadith}s are narrated from one individual to another:
    \begin{enumerate}
    \item (\emph{hadits}) → (\emph{cerita}) (confidence 0.408115, lift 4.990678),
    \item (\emph{cerita}) → (\emph{kepada}) (confidence 0.368237, lift 3.654730)
    \end{enumerate}

   Although the confidence values are lower compared to the other rules, the lift values remain high. This indicates that the narrative structures such as ''was narrated to…`` or ''comes from the story of…`` frequently appear in \emph{hadith}s and represent elements of the \emph{sanad} (chain of narrators) and \emph{matan} (content of the \emph{hadith}). In other words, the word-association approach is capable of revealing the pattern of \emph{hadith} transmission without requiring explicit structural annotation.

   The knowledge discovery that emerges here is that the structure of the \emph{sanad} in \emph{hadith}s can be indirectly identified through word-association patterns such as ''story`` and ''to``, which function as connectors between narrators. This opens opportunities for developing systems that can automatically identify patterns of transmission, even when they are not explicitly labeled as \emph{sanad}.

\end{enumerate}

\subsubsection{Validasi dan Implikasi}

Validation of the thematic clustering results in this study was carried out by examining the association metrics quantitatively and through domain-knowledge evaluation of the semantic meaning and scholarly context of \emph{hadith} studies. The quantitative approach was used to ensure that the co-occurring word patterns exhibit significant association strength, while semantic validation was performed to assess the relevance of these patterns to the conceptual structures known in Islamic studies. The combination of both provides an objective as well as contextual foundation for evaluating the validity of the thematic clusters.

Analysis of the association results reveals thematic relationships that are consistent with the semantic structure of \emph{hadith}. Several rules reflect strong connections within the theme of worship, such as the relationship between ritual components and religious practices, while other rule groups display close relations with the contexts of revelation and \emph{hadith} transmission. These connections indicate that word co-occurrence patterns in \emph{hadith} texts are not merely linguistic phenomena, but representations of a network of meanings that reflect the structure of religious knowledge.

Semantic validation was performed by re-examining the clustered results against traditional \emph{hadith} structures, including the \emph{sanad}, \emph{matan}, and the context of message transmission. The findings show a high degree of alignment between the formed thematic groups and the common narrative structure of \emph{hadith}, such as transmission patterns, worship themes, and revelation accounts. Thus, the resulting clusters can be considered structurally and conceptually valid, and relevant within the context of Islamic studies, particularly in the fields of \emph{hadith} and fiqh.

Implicitly, these results indicate that an unsupervised approach such as Apriori can be used to:

\begin{enumerate} 
\item Building automatic thematic annotations for religious texts based on word co-occurrence.

\item Revealing the structure of meaning within \emph{hadith} texts without the need for manual supervision or domain-based annotation.

\item Serving as a foundation for developing a \emph{hadith} knowledge graph, in which relationships between entities (such as ''\emph{shalat},`` ''\emph{rakaat},`` ''\emph{malam},`` ''\emph{turun}``) are automatically constructed from data co-occurrence.

\item Supporting the development of digital Islamic studies systems such as thematic \emph{hadith} search engines, meaning-based content recommendation systems, or digital \emph{sanad}-tracking models.
\end{enumerate}

These findings reinforce the argument that validation based on a combination of quantitative metrics and Islamic domain knowledge can produce an analysis that balances data objectivity with depth of meaning. Such an approach allows the interpretation of results to move beyond the statistical level and toward a conceptual understanding that is relevant to the theological values and epistemological structure of Islam. Thus, data mining–based research not only contributes to the technical aspects of text analysis but also enriches scientific methodologies in contemporary Islamic studies.

This dual-validation approach also opens space for cross-disciplinary collaboration between data scientists and Islamic scholars or \emph{hadith} researchers. Such collaboration is essential to ensure that the interpretation of analytical results remains aligned with the context of Islamic scholarship without losing technological accuracy. \emph{hadith} experts can evaluate the meaning and theological relevance of the associations found, while data science experts ensure statistical validity and algorithmic efficiency. This synergy has the potential to create a new research model that is data-driven yet firmly rooted in traditional scholarly authority.

The validation results indicate a significant potential for the development of a more adaptive and intelligent digital ecosystem for Islamic knowledge. With the ability to identify thematic patterns automatically and verify their validity, data mining–based systems can serve as a foundation for more structured and traceable digitalization of religious sources. Such models may be applied in Islamic learning platforms, digital fatwa systems, or thematic analysis of the Qur’an and \emph{hadith}, thereby expanding public access to religious knowledge that is authentic, contextual, and supported by scientific evidence.

The integration of computational methods with classical Islamic studies encourages the emergence of new analytical frameworks that can respond to contemporary research challenges. As religious texts continue to be explored within increasingly digitized environments, the availability of algorithmic tools enables researchers to revisit long-standing discussions with fresh empirical evidence. This does not replace traditional interpretative methodologies; rather, it complements them by offering alternative perspectives that can confirm, challenge, or refine existing scholarly conclusions. In this sense, data-driven validation becomes a catalyst for methodological renewal while maintaining reverence for authoritative textual traditions.

The growing relevance of digital methodologies in Islamic scholarship highlights the importance of developing standardized evaluation protocols to ensure the reliability of computational findings. Establishing guidelines for integrating statistical metrics with domain-specific knowledge will help maintain consistency across studies and reduce interpretative bias. Such protocols may include criteria for selecting threshold values in association mining, frameworks for semantic validation by experts, and mechanisms for documenting interpretative justifications. By institutionalizing these standards, researchers can create a more transparent and reproducible research environment that strengthens the academic credibility and long-term sustainability of data-driven Islamic studies.

\subsection{Discussion}

The application of the Apriori algorithm to the corpus of \emph{hadith} texts in this study opens a new perspective on non-manual approaches to thematic interpretation of Islamic sources. The results show that word associations within the text can serve as strong representations of dominant themes that are not always explicitly stated but recur consistently within the semantic structure of \emph{hadith} narratives.

Supported by high confidence levels and substantial lift values, indicate a strong semantic connection between the concept of \emph{rakaat} and the practice of \emph{shalat}, thereby reinforcing the construction of meaning within ritual contexts. A similar pattern appears in the association between references to \emph{ayat} and the process of \emph{turun}, as well as the link between the act of \emph{baca} and the text of the \emph{ayat}. Interestingly, these relationships form an implicit depiction of the revelation process without explicitly mentioning terms such as \emph{wahyu} or \emph{Jibril}. This demonstrates that word associations can uncover latent meaning structures embedded within religious narratives.

These findings illustrate that unsupervised thematic grouping using Apriori is not merely a data exploration method but also a relevant instrument for detecting latent meaning structures in religious texts. Compared to traditional methods that require manual annotation or expert labeling, this approach offers scalability and efficiency while maintaining interpretive consistency through quantitative metrics such as support, confidence, and lift.

\subsubsection{Contribution to Digital Islamic Studies}

From the perspective of Islamic scholarship, this research contributes to a methodological shift from manual philological approaches toward data-driven methods. Findings such as (\emph{hadits}) → (\emph{cerita}) and (\emph{cerita}) → (\emph{kepada}) provide a pathway for systems to understand patterns of \emph{hadith} transmission (\emph{sanad}) without requiring structural formats or predefined tags. This represents a form of discovery-based knowledge representation, where meaning is derived from data structures rather than solely from editorial structures.

Furthermore, the successful identification of themes such as worship, revelation, and \emph{hadith} transmission through word associations brings \emph{hadith} studies closer to systematic text-based methodologies, without diminishing the conceptual depth embedded within the content. This demonstrates that the approach aligns with the needs of modern society in accessing religious sources that are accurate, fast, and logically traceable.

A data-driven approach in \emph{hadith} studies also opens opportunities for integrating Islamic sciences with computer science, especially in the fields of text mining and natural language processing (NLP). With algorithmic analysis of \emph{hadith} corpora, researchers can trace semantic relationships between words or phrases that were previously difficult to identify using manual methods. This integration enriches Islamic scholarship by adding quantitative and analytical dimensions capable of uncovering hidden patterns within religious texts.

The application of this method further strengthens the validity of \emph{hadith} research by providing an empirical basis for hypothesis testing. For example, the consistent appearance of certain word patterns within specific chains of transmitters (\emph{sanad}) can serve as indicators of authenticity or highlight the linguistic tendencies of particular narrators. Thus, the research is not only descriptive but also contributes verifiable insights to the Islamic scholarly tradition, consistent with the classical methods developed by \emph{hadith} scholars in \emph{sanad} and \emph{matan} criticism.

This methodological transformation affirms that Islamic scholarship has significant potential to adapt to modern scientific paradigms. Using data as the foundation for analysis is not a form of secularization of knowledge, but rather an innovative step to strengthen the epistemology of Islam so that it remains relevant in the digital era. When utilized appropriately, technology can deepen, structure, and broaden the study of \emph{hadith} across disciplines without compromising its spiritual value or the authenticity of its sources.

\subsubsection{Relevance to the Initial Problem}

This research originated from the urgent need to address the highly manual and inefficient methods of thematic analysis in \emph{hadith} studies. By applying the Apriori algorithm, the process of extracting themes from hundreds or even thousands of \emph{hadith}s can be performed automatically, paving the way for an automated thematic annotation system. Furthermore, the findings of this study demonstrate that the associations discovered can be validated against the knowledge structures inherent in \emph{hadith} studies, indicating that this approach is not only technical but also epistemological.

Thus, the initial goal of the research—to provide an efficient, objective, and scalable thematic system—has been achieved, while also yielding additional findings in the form of latent meaning relationship patterns that were previously undetectable through explicit approaches.

\subsubsection{Relevance to SDG 4 – Quality Education}

From a socio-educational perspective, this approach has direct implications for the Sustainable Development Goals (SDG) 4: Quality Education. The findings enable the development of more inclusive, technology-based Islamic learning systems, such as automated thematic \emph{hadith} search platforms, applications for teaching \emph{hadith} meanings in contextualized ways, or digital tools for teachers and students in Islamic boarding schools (\emph{pesantren}). By utilizing data mining techniques to provide accurate and easily accessible religious knowledge, this research bridges the classical Islamic scholarly tradition with contemporary technology.

Despite its significant potential, this approach has several limitations:

\begin{enumerate}
\item  Syntactic Context Reduction

Apriori does not take word order or grammar into account. As a result, the associations formed are at a low semantic level and do not reflect syntactic or pragmatic relationships within the \emph{hadith} texts.

\item  Possibility of Spurious Associations

Some rules, even though they meet the metric thresholds, still carry the risk of containing random associations that are conceptually irrelevant. This necessitates expert validation to ensure the quality of the meaning.

\item  Scalability Constraints

In very large \emph{hadith} datasets, the computational complexity of Apriori can become a practical bottleneck, particularly when generating a large number of rules that are difficult to analyze manually.

\end{enumerate}

With semantic validation and thematic clustering applied in this study, the majority of the rules have been shown to be conceptually sound and relevant to the meaning structures within \emph{hadith} studies.

In the future, this approach can be further developed by:
\begin{enumerate}

\item Integration with word embeddings (e.g., Word2Vec, BERT) to capture associations based on semantic vector representations.

\item Utilization of a knowledge graph to connect \emph{hadith} entities conceptually and hierarchically..

\item  Development of an automatic labeling (auto-annotation) system that can assist students and researchers in instantly accessing \emph{hadith} themes based on inter-concept relationships.
\end{enumerate}

This data-driven approach demonstrates that Islamic studies have considerable potential to transform toward more empirical and measurable methodologies. The use of algorithms such as Apriori provides a foundation for establishing an objective analytical framework for \emph{hadith} texts, allowing the knowledge obtained to be not only interpretative but also verifiable through data. This transformation strengthens the position of Islamic studies in the digital era by balancing scientific accuracy with the depth of theological meaning.

The application of this method also contributes to the modernization of Islamic education curricula. \emph{hadith} content can be presented interactively and contextually through visualizations of thematic relationships, concept networks, or semantic maps between \emph{hadith}s. Such innovations have the potential to enhance digital literacy among students in \emph{pesantren} as well as Islamic academics, while also broadening access to authentic sources of knowledge that have traditionally been dispersed across thousands of classical texts. Integrating text analysis technology with classical scholarship can reinforce scientific thinking in understanding Islamic teachings in a rational and profound manner.

Further development opportunities for this research lie in building a collaborative and open digital ecosystem for Islamic knowledge. By combining data mining, natural language processing, and validation within the domain of Islamic scholarship, the resulting intelligent system can support interdisciplinary research—ranging from \emph{hadith} studies and \emph{tafsir} to Islamic history. Such development not only promotes research efficiency but also expands public access to organized, traceable Islamic knowledge while maintaining the authenticity of its scholarly sources.

\section{Conclusion} \label{sec:conclusion}

This study demonstrates that an unsupervised approach based on the Apriori algorithm is capable of systematically and objectively extracting thematic knowledge from a corpus of \emph{hadith} texts. Through word association analysis, the system successfully forms rules reflecting significant semantic relationships, such as the connection between the concepts of ''\emph{shalat}`` and ''\emph{rakaat},`` ''\emph{malam},`` and ''\emph{diri},`` as well as the relationship among ''\emph{ayat},`` ''\emph{turun},`` and ''\emph{baca},`` which denote narratives of revelation. These findings confirm that semantic relationships can be detected through co-occurrence patterns, even without manual annotation or expert supervision.

The main contributions of this research lie in two aspects. First, methodologically, this approach fills a gap among previous studies that still relied on manual labeling or supervised classification methods. Second, in terms of content, the obtained associations approximate the conceptual structures known in \emph{hadith} studies, such as \emph{matan}, \emph{\emph{sanad}}, and the contexts of worship and prophethood, with strong semantic validation through lift and confidence metrics.

Nevertheless, there are several limitations to note. The Apriori approach does not consider syntactic structure or sentence order, and thus cannot fully capture the complexity of meaning within the construction of \emph{hadith} sentences. In addition, the potential formation of spurious or coincidental associations still requires expert involvement during the interpretive stage.

Practically, this research opens the way for developing applications based on thematic understanding of \emph{hadith}s, ranging from automatic annotation systems, theme-based navigation, to the construction of Islamic knowledge graphs. In the context of Islamic education and digital literacy, this approach supports inclusive, standardized, and efficient learning, in line with Sustainable Development Goal (SDG) 4.

For future development, this research can be expanded through integration with semantic representations based on word embeddings, topic modeling, or graph-based learning, enabling the exploration of meaning to encompass more complex syntactic, pragmatic, and historical contexts within \emph{hadith} texts.

\section*{Acknowledgment}
The author's wishes to acknowledge the Informatics Department UIN Sunan Gunung Djati Bandung, which partially supports this research work.

%\begin{thebibliography}{00}
\bibliographystyle{./IEEEtran}
\bibliography{./IEEEabrv,./IEEEkelompok1}

@inproceedings{ramponi2020unsupervised,
  author    = {A. Ramponi and B. Plank},
  title     = {Neural Unsupervised Domain Adaptation in NLP---A Survey},
  booktitle = {Proceedings of the 28th International Conference on Computational Linguistics},
  address   = {Stroudsburg, PA, USA},
  publisher = {International Committee on Computational Linguistics},
  year      = {2020},
  pages     = {6838--6855},
  doi       = {10.18653/v1/2020.coling-main.603}
}

@article{alfianzah2020apriori,
  author    = {R. Alfianzah and R. I. Handayani and M. Murniyati},
  title     = {Implementation of Apriori Algorithm Data Mining for Increase Sales},
  journal   = {SinkrOn},
  volume    = {5},
  number    = {1},
  pages     = {17},
  year      = {2020},
  doi       = {10.33395/sinkron.v5i1.10587},
  month     = {oct}
}

@article{hibnastiar2025apriori,
  author    = {N. A. Hibnastiar and A. F. Setiawan and E. H. Susanto},
  title     = {Penerapan Algoritma Apriori dalam Menentukan Rekomendasi Paket Produk},
  journal   = {MALCOM: Indonesian Journal of Machine Learning and Computer Science},
  volume    = {5},
  number    = {1},
  pages     = {321--331},
  year      = {2025},
  doi       = {10.57152/malcom.v5i1.1782},
  month     = {jan}
}

@article{do2020humanrights,
  author    = {D.-N.-M. Do and L.-K. Hoang and C.-M. Le and T. Tran},
  title     = {A Human Rights-Based Approach in Implementing Sustainable Development Goal 4 (Quality Education) for Ethnic Minorities in Vietnam},
  journal   = {Sustainability},
  volume    = {12},
  number    = {10},
  pages     = {4179},
  year      = {2020},
  doi       = {10.3390/su12104179},
  month     = {may}
}

@article{ferguson2020sdg,
  author    = {T. Ferguson and C. G. Roofe},
  title     = {SDG 4 in higher education: challenges and opportunities},
  journal   = {International Journal of Sustainability in Higher Education},
  volume    = {21},
  number    = {5},
  pages     = {959--975},
  year      = {2020},
  month     = {jun},
  doi       = {10.1108/IJSHE-12-2019-0353}
}

@incollection{sayed2020quality,
  author    = {Y. Sayed and K. Moriarty},
  title     = {SDG 4 and the ‘Education Quality Turn’},
  booktitle = {Grading Goal Four},
  publisher = {BRILL},
  year      = {2020},
  pages     = {194--213},
  doi       = {10.1163/9789004430365_009}
}

@incollection{franco2020sdg,
  author    = {I. B. Franco and E. Derbyshire},
  title     = {SDG 4 Quality Education},
  booktitle = {Envisioning Futures for Environmental and Sustainability Education},
  year      = {2020},
  pages     = {57--68},
  doi       = {10.1007/978-981-32-9927-6_5}
}

@incollection{raja2021ai,
  author    = {R. A. Raja and N. Yuvaraj and N. V. Kousik},
  title     = {Analyses on Artificial Intelligence Framework to Detect Crime Pattern},
  booktitle = {Intelligent Data Analytics for Terror Threat Prediction},
  publisher = {Wiley},
  year      = {2021},
  pages     = {119--132},
  doi       = {10.1002/9781119711629.ch6}
}

@inproceedings{dharshinni2019kmeans,
  author    = {N. P. Dharshinni and F. Azmi and I. Fawwaz and A. M. Husein and S. D. Siregar},
  title     = {Analysis of Accuracy K-Means and Apriori Algorithms for Patient Data Clusters},
  booktitle = {Journal of Physics: Conference Series},
  volume    = {1230},
  number    = {1},
  pages     = {012020},
  year      = {2019},
  month     = {jul},
  doi       = {10.1088/1742-6596/1230/1/012020}
}

@inproceedings{inproceedings,
author = {Noor, Umar},
year = {2020},
month = {10},
pages = {711-719},
title = {The Revival Of Hadith Study In Modern Time},
doi = {10.15405/epsbs.2020.10.02.66}
}

@article{shao2020conceptmap,
  author    = {Z. Shao and Y. Li and X. Wang and X. Zhao and Y. Guo},
  title     = {Research on a New Automatic Generation Algorithm of Concept Map Based on Text Analysis and Association Rules Mining},
  journal   = {Journal of Ambient Intelligence and Humanized Computing},
  volume    = {11},
  number    = {2},
  pages     = {539--551},
  year      = {2020},
  month     = feb,
  doi       = {10.1007/s12652-018-0934-9}
}

@article{tarigan2022implementasi,
  title={Implementasi data mining menggunakan algoritma apriori dalam menentukan persediaan barang: Studi kasus: Toko sinar harahap},
  author={Tarigan, Putri Mai Sarah and Hardinata, Jaya Tata and Qurniawan, Hendry and Safii, Muhammad and Winanjaya, Riki},
  journal={Jurnal Janitra Informatika dan Sistem Informasi},
  volume={2},
  number={1},
  pages={9--19},
  year={2022}
}

@article{nurmalasari2022implementasi,
  title={Implementasi Ekstraksi Fitur untuk Pengelompokan Dokumen Proposal Menggunakan Algoritma Na{\"\i}ve Bayes},
  author={Nurmalasari, Dini and Yuliantoro, Heri Ribut},
  journal={Jurnal Komputer Terapan},
  volume={8},
  number={1},
  pages={473014},
  year={2022},
  publisher={Politeknik Caltex Riau}
}

@article{sinaga2020unsupervised,
  title={Unsupervised K-means clustering algorithm},
  author={Sinaga, Kristina P and Yang, Miin-Shen},
  journal={IEEE access},
  volume={8},
  pages={80716--80727},
  year={2020},
  publisher={IEEE}
}

@article{az2024implementasi,
  title={IMPLEMENTASI ALGORITMA APRIORI PADA ANALISIS POLA PENJUALAN SEPATU},
  author={Az-zahra, Nabila Sofia and Fatah, Zaehol},
  journal={Jurnal Riset Teknik Komputer},
  volume={1},
  number={4},
  pages={73--79},
  year={2024}
}

@article{yudhatama2024penerapan,
  title={Penerapan Metode Association Rule Menggunakan Apriori untuk Rekomendasi Penyusunan Rak Buku di Perpustakaan Universitas Muhammadiyah Bengkulu},
  author={Yudhatama, M Rafli and Wijaya, Ardi and Abdullah, Dedy and Sonita, Anisya},
  journal={Jurnal Multidisiplin Dehasen (MUDE)},
  volume={3},
  number={3},
  pages={163--168},
  year={2024}
}

@data{FK2/GWSEWB_2022,
author = {Adiwijaya, Adiwijaya and Said Al Faraby and Mohamad Syahrul Mubarok and Mahendra Dwifebri Purbolaksono},
publisher = {Telkom University Dataverse},
title = {{Indonesian Translation of the Hadith of Bukhari (Single-label)}},
UNF = {UNF:6:P7HgGoiV1wK6ZSrMaSQSnA==},
year = {2022},
version = {DRAFT VERSION},
doi = {10.34820/FK2/GWSEWB},
url = {https://doi.org/10.34820/FK2/GWSEWB}
}

@article{alinafiah2024implementasi,
  title={Implementasi Impementasi Data Mining Dalam Pengelolaan Stok Obat Menggunakan Metode K-Means Clustering dan Asossociation Rules Apriori: Analisis Pola Pembelian dan Hubungan Antar Obat dalam Pengelolaan Stok menggunakan K-Means Clustering dan Association Rules Apriori},
  author={Alinafiah, Andika Muhammad and Octariadi, Barry Ceasar and Sucipto, Sucipto},
  journal={Jurnal Informatika Polinema},
  volume={10},
  number={4},
  pages={551--558},
  year={2024}
}

@incollection{lischner2014case,
  title={Case-folding},
  author={Lischner, Ray},
  booktitle={Exploring C++ 11: Problems and Solutions Handbook},
  pages={111--113},
  year={2014},
  publisher={Springer}
}

@inproceedings{rosid2020improving,
  title={Improving text preprocessing for student complaint document classification using sastrawi},
  author={Rosid, Mochamad Alfan and Fitrani, Arif Senja and Astutik, Ika Ratna Indra and Mulloh, Nasrudin Iqrok and Gozali, Haris Ahmad},
  booktitle={IOP Conference Series: Materials Science and Engineering},
  volume={874},
  number={1},
  pages={012017},
  year={2020},
  organization={IOP Publishing}
}

@inproceedings{bougar2019addressing,
  title={Addressing Stemming Algorithm for Arabic Text Using Spark Over Hadoop},
  author={Bougar, Marieme and Ziyati, El Houssaine},
  booktitle={International Conference on Advanced Intelligent Systems for Sustainable Development},
  pages={74--82},
  year={2019},
  organization={Springer}
}

@article{nurhidayanti2022apriori,
  author    = {Dina Nurhidayanti and Ika Kurniawati},
  title     = {Implementasi Algoritma Apriori Dalam Menemukan Association Rules Pada Persediaan Sparepart Motor},
  journal   = {Innovation in Research of Informatics (INNOVATICS)},
  volume    = {4},
  number    = {2},
  pages     = {62--67},
  year      = {2022},
  doi       = {10.37058/innovatics.v4i2.5300},
  url       = {https://jurnal.unsil.ac.id/index.php/innovatics/article/view/5300},
  publisher = {Universitas Siliwangi},
  note      = {Accessed: 2025-11-05}
}

@article{Doskarayev2023,
title = {Development of Computer Vision-enabled Augmented Reality Games to Increase Motivation for Sports},
journal = {International Journal of Advanced Computer Science and Applications},
doi = {10.14569/IJACSA.2023.0140428},
url = {http://dx.doi.org/10.14569/IJACSA.2023.0140428},
year = {2023},
publisher = {The Science and Information Organization},
volume = {14},
number = {4},
author = {Bauyrzhan Doskarayev and Nurlan Omarov and Bakhytzhan Omarov and Zhuldyz Ismagulova and Zhadra Kozhamkulova and Elmira Nurlybaeva and Galiya Kasimova}
}

@article{FatinaWahid2021,
title = {A Novel Threshold based Method for Vessel Intensity Detection and Extraction from Retinal Images},
journal = {International Journal of Advanced Computer Science and Applications},
doi = {10.14569/IJACSA.2021.0120663},
url = {http://dx.doi.org/10.14569/IJACSA.2021.0120663},
year = {2021},
publisher = {The Science and Information Organization},
volume = {12},
number = {6},
author = {Farha Fatina Wahid and Sugandhi K and Raju G and Debabrata Swain and Biswaranjan Acharya and Manas Ranjan Pradhan}
}

@article{Alhasan2021,
title = {Evaluation of Data Center Network Security based on Next-Generation Firewall},
journal = {International Journal of Advanced Computer Science and Applications},
doi = {10.14569/IJACSA.2021.0120958},
url = {http://dx.doi.org/10.14569/IJACSA.2021.0120958},
year = {2021},
publisher = {The Science and Information Organization},
volume = {12},
number = {9},
author = {Andi Jehan Alhasan and Nico Surantha}
}

@INPROCEEDINGS{7577586,
  author={Zulfikar, Wildan Budiawan and Wahana, Agung and Uriawan, Wisnu and Lukman, Nur},
  booktitle={2016 4th International Conference on Cyber and IT Service Management}, 
  title={Implementation of association rules with apriori algorithm for increasing the quality of promotion}, 
  year={2016},
  volume={},
  number={},
  pages={1-5},
  doi={10.1109/CITSM.2016.7577586}}

@INPROCEEDINGS{10903699,
  author={Asy'ari, Ahmad Hasyim and Muzakki, Mohammad Haris and Hanafi, M.},
  booktitle={2024 6th International Conference on Cybernetics and Intelligent System (ICORIS)}, 
  title={Clusterization Model of Hadith Topic in Bukhari Muslim Hadith using BERT Algorithm}, 
  year={2024},
  volume={},
  number={},
  pages={1-6},
  doi={10.1109/ICORIS63540.2024.10903699}}

%\end{thebibliography}

\end{document}